\pdfoutput=1

\documentclass[11pt]{article}

\usepackage{ACL2023}

\usepackage{times}
\usepackage{latexsym}

\usepackage[T1]{fontenc}

\usepackage[utf8]{inputenc}

\usepackage{microtype}

\usepackage{inconsolata}
\usepackage{graphicx}
\usepackage{tikz}
\usepackage{adjustbox}
\usepackage[draft]{todo}
\usepackage[capitalise, noabbrev]{cleveref}

\usepackage[normalem]{ulem} 
\usepackage{xcolor}

\title{Assessing Word Importance Using Models Trained for Semantic Tasks}

\author{Dávid Javorský$^1$ \and Ondřej Bojar$^1$ \and François Yvon$^2$
  \\ \\
  $^{1}$Charles University, Faculty of Mathematics and Physics, Prague, Czechia\\
  $^2$Sorbonne Université, CNRS, ISIR, Paris, France\\
  \texttt{\{javorsky,bojar\}@ufal.mff.cuni.cz}~~~\texttt{francois.yvon@cnrs.fr}}

\begin{document}
\maketitle
\begin{abstract}

Many NLP tasks require to automatically identify the most significant words in a text. In this work, we derive word significance from models trained to solve semantic task: Natural Language Inference and Paraphrase Identification. Using an attribution method aimed to explain the predictions of these models, we derive importance scores for each input token. We evaluate their relevance using a so-called cross-task evaluation: Analyzing the performance of one model on an input masked according to the other model's weight, we show that our method is robust with respect to the choice of the initial task. Additionally, we investigate the scores from the syntax point of view and observe interesting patterns, e.g.\ words closer to the root of a syntactic tree receive higher importance scores. Altogether, these observations suggest that our method can be used to identify important words in sentences without any explicit word importance labeling in training.

\end{abstract}

\section{Introduction}

The ability to decide which words in a sentence are semantically important plays a crucial role in various areas of NLP (e.g. compression, paraphrasing, summarization, keyword identification). One way to compute (semantic) word significance for compression purposes is to rely on syntactic patterns, using Integer Linear Programming techniques to combine several sources of information \cite{clarke-lapata-2006-constraint, filippova-strube-2008-dependency}. \citet{xu-grishman-2009-parse} exploit the same cues, with significance score computed as a mixture of TF-IDF and surface syntactic cues. A similar approach estimates word importance for summarization \citep{hong-nenkova-2014-improving} or learns these significance scores from word embeddings \citep{schakel2015measuring,sheikh-etal-2016-learning}.

\begin{figure}[t]
    \centering
    \includegraphics[width=0.48\textwidth]{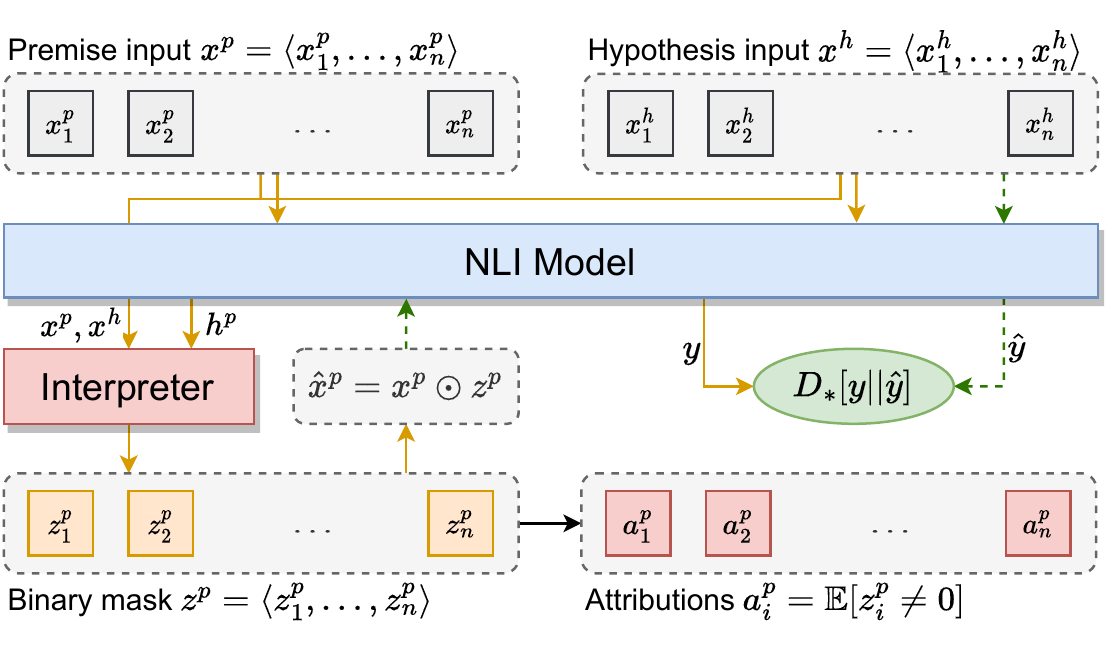}
    \caption{The first pass (yellow plain arrows): A premise and hypothesis are passed to the NLI model. The interpreter takes both text inputs $x^p$, $x^h$, and hidden states $h^p$ of the NLI model's encoder. It generates a binary mask $z^p$ which is used to mask $x^p$, resulting in $\hat{x}^p$. The second pass (green dashed arrows): $\hat{x}^p$ is passed to the NLI model together with the original hypothesis. The divergence $D_*$ minimizes the difference between predicted distributions $y$ and $\hat{y}$ of these two passes.}
    \label{fig:pipeline}
\end{figure}

Significance scores are also useful in an entirely different context, that of explaining the decisions of Deep Neural Networks (DNNs). This includes investigating and interpreting hidden  representations via auxiliary probing tasks \cite{adi2016fine,conneau-etal-2018-cram}; quantifying the importance of input words in the decisions computed by DNNs in terms of analyzing attention patterns \cite{clark-etal-2019-bert}; or using attribution methods based on  attention \cite{vashishth2019attention}, back-propagation \cite{sundararajan2017axiomatic} or perturbation techniques \citep{pmlr-v97-guan19a,schulz2020restricting}. Along these lines, \citet{deyoung-etal-2020-eraser} present a benchmark for evaluating the quality of model-generated rationals compared to human rationals.

In this study, we propose to use such techniques to compute semantic significance scores in an innovative way. We demand the scores to have these intuitive properties: (a) Content words are more important than function words; (b) Scores are context-dependent; (c) Removing low-score words minimally changes the sentence meaning. For this, we train models for two semantic tasks, Natural Language Inference and Paraphrase Identification, and use the attribution approach of \citet{de-cao-etal-2020-decisions} to explain the models' predictions. We evaluate the relevance of scores using the so-called \emph{cross-task evaluation}: Analyzing the performance of one model on an input masked according to the other model's weights. We show that our method is robust with respect to the choice of the initial task and fulfills all our requirements. Additionally, hinting at the fact that trained hidden representations encode a substantial amount of linguistic information about morphology \citep{belinkov-etal-2017-neural}, syntax \citep{clark-etal-2019-bert,hewitt-manning-2019-structural}, or both \citep{peters-etal-2018-dissecting}, we also analyze the correlations of our scores with syntactic patterns.

\section{Method \label{sec:method}}

\begin{figure}[t]
    \centering
    \includegraphics[width=0.2\textwidth]{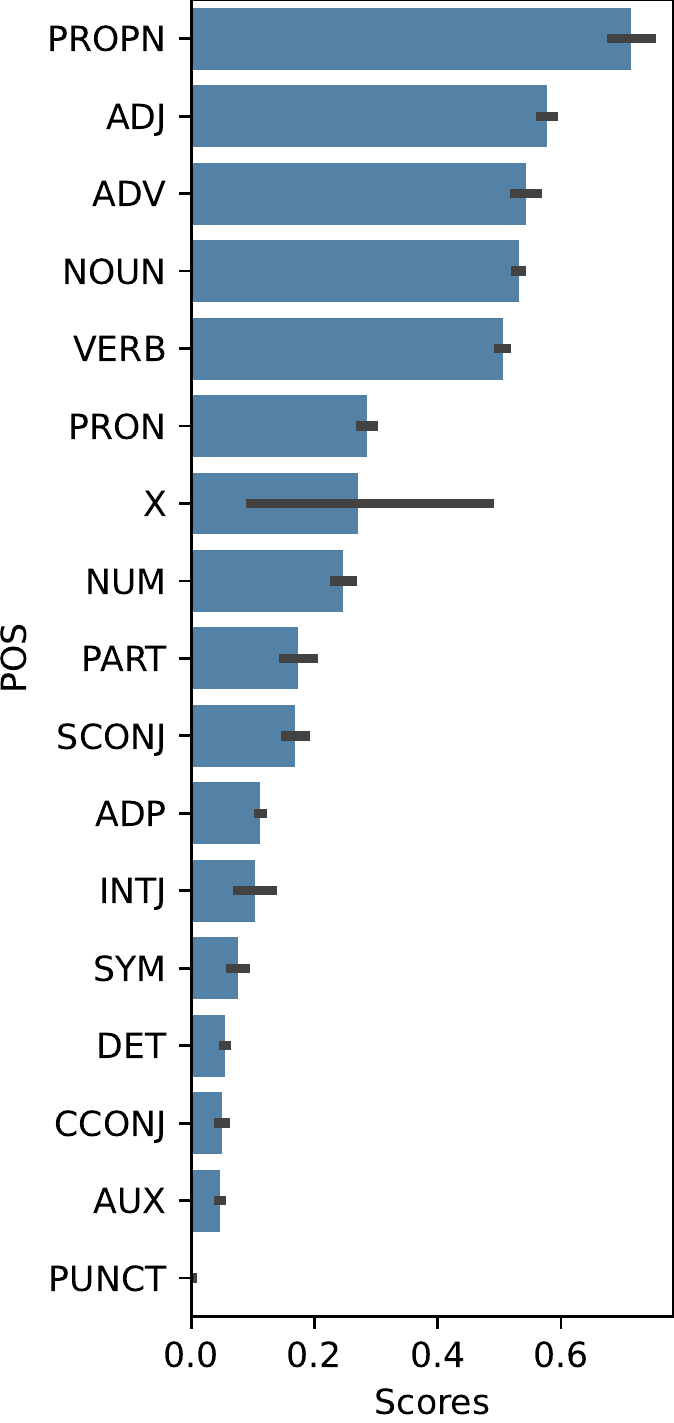}
    \includegraphics[width=0.2\textwidth]{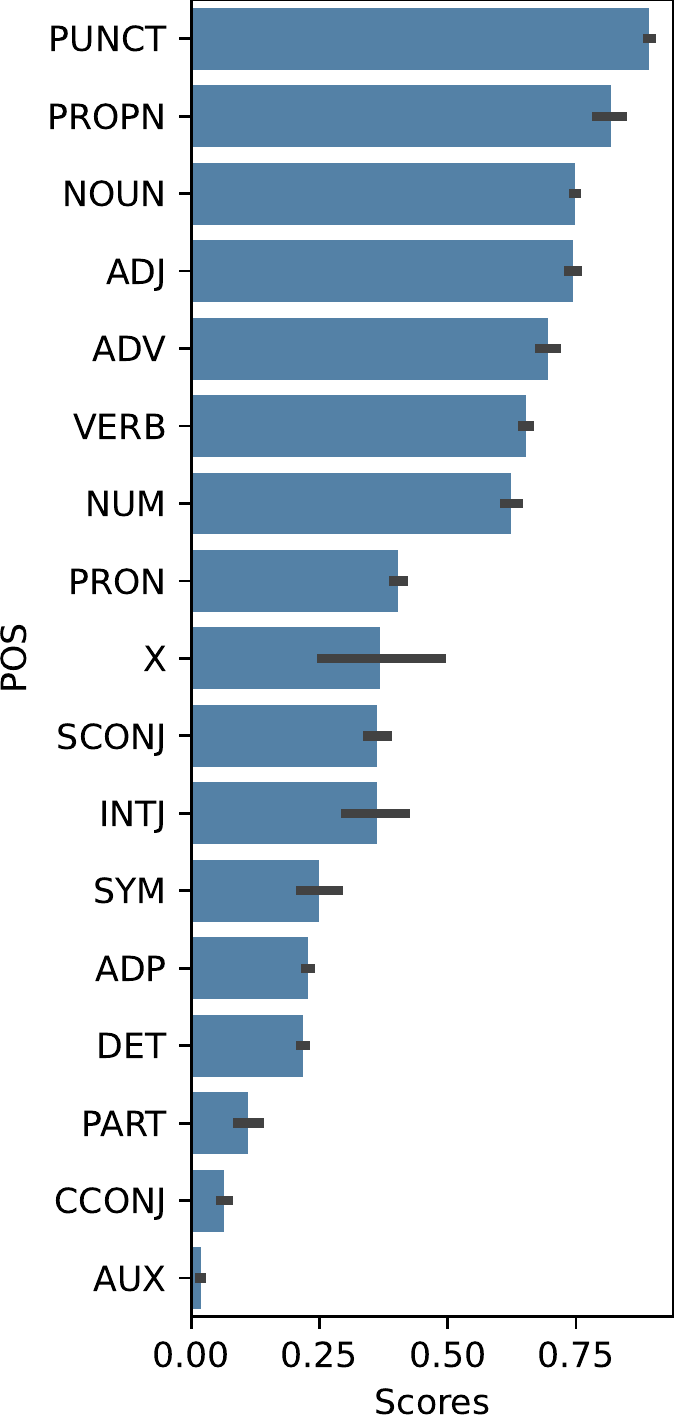}
    \caption{Average scores for each POS category for the NLI model (left) and PI model (right).}
    \label{fig:pos_analysis}
\end{figure}

We assume that sentence-level word significance (or word importance) is assessed by the amount of  contribution to the overall meaning of the sentence. This means that removing low-scored word should only slightly change the sentence meaning.

The method we explore to compute significance score repurposes attribution techniques originally introduced to explain the predictions of a DNN trained for a specific task. Attribution methods typically compute sentence level scores for each input word, identifying the ones that contribute most to the decision. By explicitly targeting semantic prediction tasks, we hope to extract attribution scores that correlate well with semantic significance.

Our significance scoring procedure thus consists of two main components: an underlying model and an interpreter. The underlying model is trained to solve a semantic task. We select two tasks: Natural Language Inference (NLI) --- classifying the relationship of a premise--hypothesis pair into entailment, neutrality or contradiction --- and Paraphrase Identification (PI) --- determining whether a pair of sentences have the same meaning.

The interpreter relies on the attribution method proposed by \citet{de-cao-etal-2020-decisions}, seeking to mask the largest possible number of words in a sentence, while at the same time preserving the underlying model's decision obtained from the full sentence pair. The interpreter thus minimizes a loss function comprising two terms: an~$L_0$ term, on the one hand, forces the interpreter to maximize the number of masked elements, and a divergence term $D_*$, on the other hand, aims to diminish the difference between the predictions of the underlying model when given (a) the original input or (b) the masked input.

We take the outputs of the interpreter, i.e.\ the attribution scores, as probabilities that given words are not masked. Following \citet{de-cao-etal-2020-decisions}, these probabilities are computed assuming an underlying Hard Concrete distribution on the closed interval $[0,1]$, which assigns a non-zero probability to extreme values (0 and 1) (Fig.~9, \citealp{de-cao-etal-2020-decisions}). During interpreter training, a reparametrization trick is used (so that the gradient can be propagated backwards) to estimate its parameters. Given the Hard Concrete distribution output, the attribution score for a token expresses the expectation of sampling a non-zero value, meaning that the token should be masked (Section~2, Stochastic masks, \citealp{de-cao-etal-2020-decisions}). We illustrate the process in \cref{fig:pipeline}.

\section{Experimental Setup \label{sec:evaluation}}

\subsection{Underlying Models}

We use a custom implementation of a variant of the Transformer architecture \citep{vaswani2017attention} which comprises two encoders sharing their weights, one for each input sentence. This design choice is critical as it allows us to compute importance weights of isolated sentences, which is what we need to do in inference. We then concatenate encoder outputs into one sequence from which a fully connected layer predicts the class, inspired by Sentence-BERT \cite{reimers-gurevych-2019-sentence} architecture. See \cref{sec:appendix_models} for a discussion on the architecture choice, and for datasets, implementation and training details.

\subsection{Interpreter}
\label{sec:diffmask}

We use the attribution method introduced by \citet{de-cao-etal-2020-decisions}. The interpreter consists of classifiers, each processing hidden states of one layer and predicting the probability whether to keep or discard input tokens. See \cref{sec:appendix_interpreter} for datasets, implementation and training details.\footnote{Our source code with the license specification is available at \url{https://github.com/J4VORSKY/word-importance}}

\section{Analysis}

\begin{table*}[t!]
    \centering
    \small
    \setlength{\tabcolsep}{3pt}
    \scalebox{0.9}{%
    \begin{tabular}{c|ccccccccccc}
               & \multicolumn{10}{c}{\bf PI Model performance}    \\\hline\hline
      & \bf 0\%    & \bf 10\%       & \bf 20\%  & \bf 30\%    & \bf 40\%       & \bf 50\% & \bf 60\% & \bf 70\% & \bf 80\% & \bf 90\% & \bf 100\%    \\\hline
    0\% & \it 85.1$\uparrow$0.0 & \it 84.7$\uparrow$0.7 &  84.5$\uparrow$4.6 &  83.0$\uparrow$6.8 &  80.9$\uparrow$9.1 &  77.7$\uparrow$12.2 &  74.3$\uparrow$12.9 &  69.3$\uparrow$10.6 &  62.6$\uparrow$7.3 &  56.0$\uparrow$4.0 & \it 50.0$\uparrow$0.0 \\
    10\% & \it 84.7$\uparrow$0.9 &  84.7$\uparrow$2.0 &  84.4$\uparrow$5.7 &  82.8$\uparrow$7.6 &  81.0$\uparrow$9.9 &  77.8$\uparrow$12.9 &  74.5$\uparrow$13.4 &  69.5$\uparrow$11.3 &  62.6$\uparrow$7.5 &  55.8$\uparrow$3.8 & \it 50.0$\uparrow$0.0 \\
    20\% &  84.2$\uparrow$4.1 &  84.2$\uparrow$5.2 &  84.3$\uparrow$8.3 &  83.0$\uparrow$10.3 &  81.5$\uparrow$12.2 &  78.4$\uparrow$14.7 &  74.9$\uparrow$14.4 &  70.1$\uparrow$12.3 &  63.0$\uparrow$8.2 &  56.2$\uparrow$4.3 & \it 50.0$\uparrow$0.0 \\
    30\% &  83.1$\uparrow$6.9 &  83.1$\uparrow$7.7 &  83.3$\uparrow$11.0 &  82.6$\uparrow$12.6 &  81.8$\uparrow$15.0 &  79.0$\uparrow$16.1 &  75.6$\uparrow$15.7 &  70.9$\uparrow$13.3 &  63.5$\uparrow$8.6 &  56.3$\uparrow$4.6 & \it 50.0$\uparrow$0.1 \\
    40\% &  80.7$\uparrow$9.9 &  80.4$\uparrow$10.4 &  81.0$\uparrow$12.7 &  81.0$\uparrow$14.0 &  80.9$\uparrow$16.1 &  78.7$\uparrow$17.9 &  75.5$\uparrow$16.1 &  71.1$\uparrow$13.7 &  64.2$\uparrow$9.9 &  56.7$\uparrow$5.0 & \it 50.0$\uparrow$0.1 \\
    50\% &  77.3$\uparrow$11.3 &  77.5$\uparrow$11.6 &  78.1$\uparrow$13.5 &  78.9$\uparrow$15.0 &  78.8$\uparrow$16.6 &  78.0$\uparrow$18.3 &  75.2$\uparrow$17.0 &  71.2$\uparrow$15.0 &  64.2$\uparrow$9.6 &  56.8$\uparrow$5.0 & \it 50.0$\uparrow$0.1 \\
    60\% &  73.6$\uparrow$11.7 &  73.9$\uparrow$12.0 &  74.4$\uparrow$13.3 &  75.9$\uparrow$15.2 &  75.3$\uparrow$16.4 &  75.9$\uparrow$17.9 &  74.4$\uparrow$17.4 &  71.2$\uparrow$15.7 &  65.3$\uparrow$11.2 &  57.1$\uparrow$5.2 & \it 49.9$\downarrow$0.2 \\
    70\% &  68.4$\uparrow$10.3 &  68.8$\uparrow$11.1 &  68.7$\uparrow$11.3 &  70.2$\uparrow$12.8 &  70.7$\uparrow$14.3 &  71.1$\uparrow$15.3 &  71.0$\uparrow$15.9 &  70.3$\uparrow$15.4 &  66.4$\uparrow$13.3 &  58.2$\uparrow$6.0 & \it 50.0$\downarrow$0.3 \\
    80\% &  62.3$\uparrow$7.3 &  62.3$\uparrow$7.5 &  62.4$\uparrow$7.6 &  63.2$\uparrow$8.7 &  63.6$\uparrow$9.3 &  64.3$\uparrow$10.4 &  64.7$\uparrow$11.1 &  65.8$\uparrow$12.6 &  67.0$\uparrow$15.0 &  59.8$\uparrow$8.2 & \it 49.7$\downarrow$0.4 \\
    90\% &  56.2$\uparrow$4.0 &  56.3$\uparrow$4.1 &  56.5$\uparrow$4.4 &  56.7$\uparrow$4.7 &  57.2$\uparrow$5.3 &  57.2$\uparrow$5.4 &  57.5$\uparrow$5.5 &  58.5$\uparrow$7.1 &  60.5$\uparrow$8.8 &  63.9$\uparrow$12.1 &  50.2$\downarrow$2.4 \\
    100\% & \it 50.0$\uparrow$0.0 & \it 50.0$\downarrow$0.0 & \it 50.0$\uparrow$0.0 & \it 50.0$\uparrow$0.1 & \it 50.0$\uparrow$0.2 & \it 50.1$\uparrow$0.1 & \it 50.0$\uparrow$0.1 & \it 50.0$\downarrow$0.1 & \it 50.1$\downarrow$0.2 & \it 50.5$\downarrow$0.5 & \it 50.0$\uparrow$0.0 \\
    \end{tabular}}
    \caption{The accuracy of the PI model when a given percentage of the least important input tokens are removed from the first sentence (rows) or the second (columns) according to the NLI model's weights. Each cell contains the model accuracy (left), difference in comparison to the randomized baseline model (right) and an arrow denoting the increase ($\uparrow$) or decrease ($\downarrow$) in performance of our model compared to the baseline. The difference of values in \textit{italics} is \textit{not} statistically significant ($p<0.01$).}
    \label{tab:valid_scores_PI}
\end{table*}

In our analysis of the predicted masks, we only consider the last-layer classifier, rescaling the values so that the lowest value and the highest value within one sentence receive the scores of zero and one, respectively. All results use the \textsc{snli} validation set.

\subsection{Content Words are More Important}
\label{sec:pos}

We first examine the scores that are assigned to content and functional words. We compute the average score for each POS tag \cite{11234/1-4758} and display the results in \cref{fig:pos_analysis}. For both models, Proper Nouns, Nouns, Pronouns, Verbs, Adjectives and Adverbs have leading scores. Determiners, Particles, Symbols, Conjunctions, Adpositions are scored lower. We observe an inconsistency of the PI model scores for Punctuation. We suppose this reflects idiosyncrasies of the PI dataset: Some items contain two sentences within one segment, and these form a paraphrase pair only when the other segment also consists of two sentences. Therefore, the PI model is more sensitive to Punctuation than expected. We also notice the estimated importance of the X category varies widely, which is expected since this category is, based on its definition, a mixture of diverse word types. Overall, these results fulfil our requirement that content words achieve higher scores than function words.

\subsection{Word Significance is Context-Dependent}

\begin{figure}[t]
    \centering
    \includegraphics[width=0.23\textwidth]{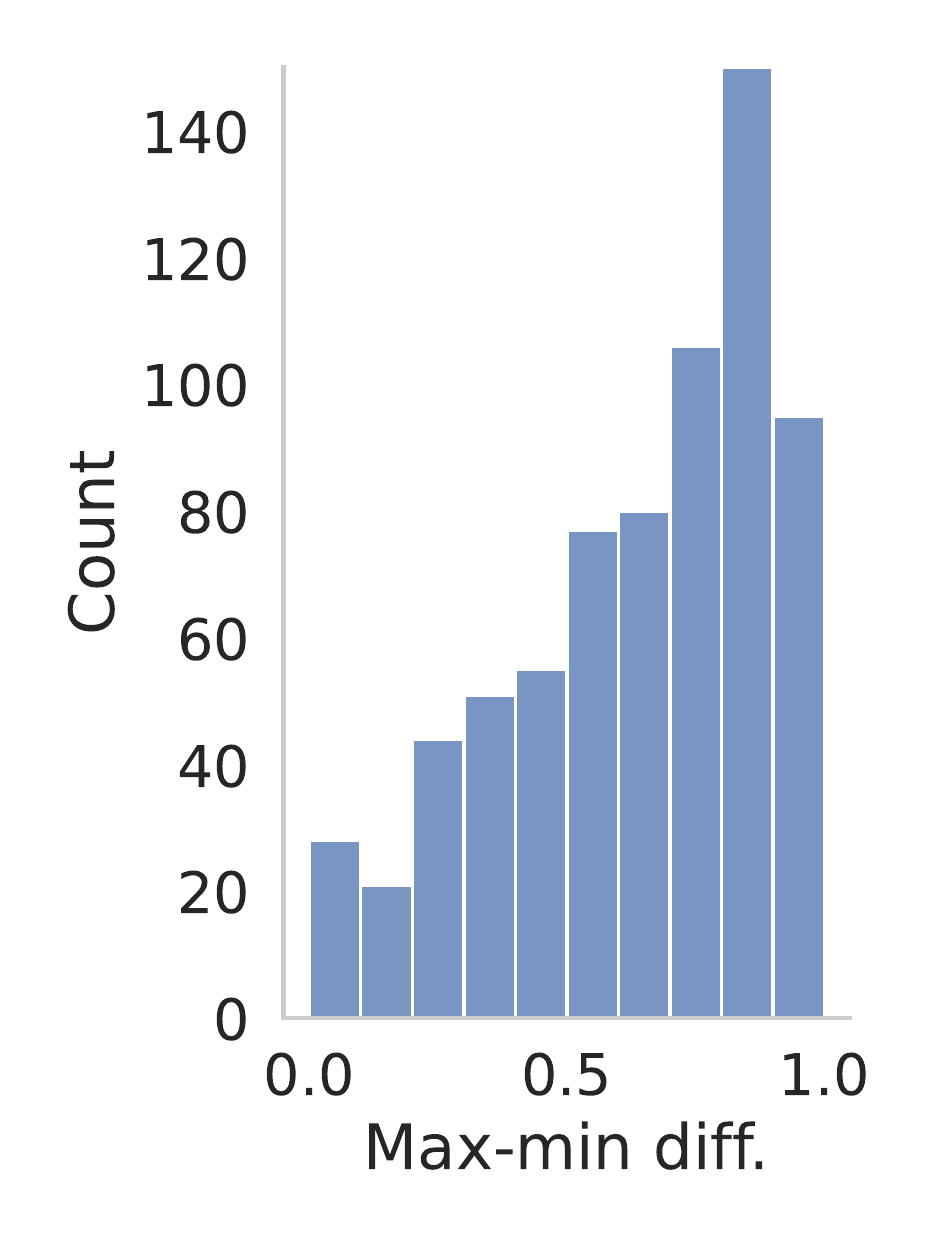}
    \includegraphics[width=0.23\textwidth]{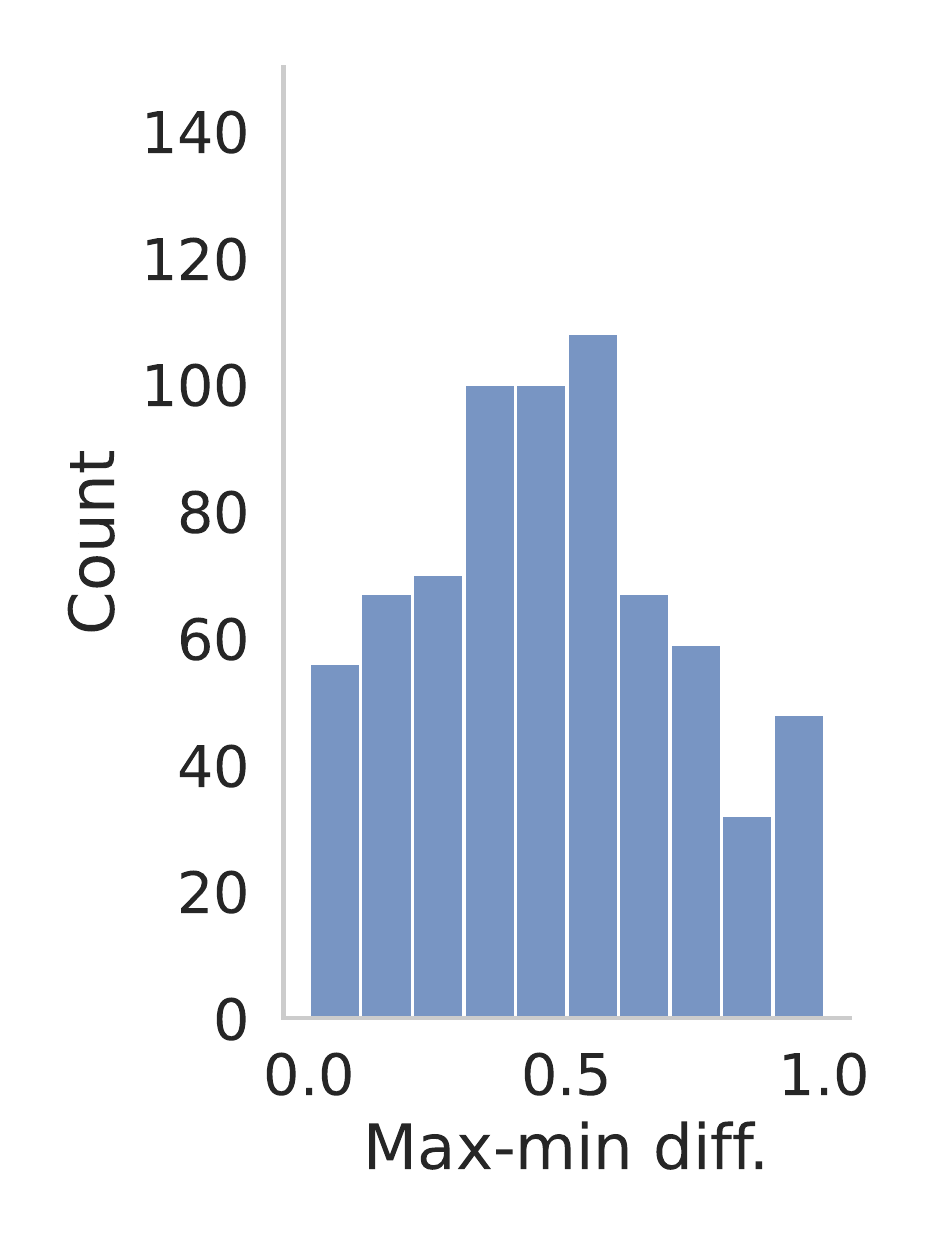}
    \caption{The NLI model (left), PI model (right) and the distribution of differences between the maximal and minimal value for each token.}
    \label{fig:max_min_diff}
\end{figure}

We then question the ability of the interpreter to generate context-dependent attributions, contrasting with purely lexical measures such as TF-IDF. To answer this question, we compute the distribution of differences between the lowest and highest scores for words having at least 100 occurrences in the training and 10 in the validation data, excluding tokens containing special characters or numerals. The full distribution is plotted in~\cref{fig:max_min_diff}.

Scores extracted from both models report increased distribution density towards larger differences, confirming that significance scores are not lexicalized, but instead strongly vary according to the context for the majority of words. The greatest difference in scores for PI model is around 0.5, the analysis of the NLI model brings this difference even more towards 1. We explain it by the nature of datasets: It is more likely that the NLI model's decision relies mostly on one or on a small group of words, especially in the case of contradictions.

\subsection{Cross-Task Evaluation}

In this section, we address the validity of importance scores. We evaluate the models using so-called \emph{cross-task evaluation}: For model A, we take its validation dataset and gradually remove a portion of the lowest scored tokens according to the interpreter of model B. We then collect the predictions of model A using the malformed inputs and compare it to a baseline where we randomly remove the same number of tokens. We evaluate both models in this setting, however, since the results for both models have similar properties, we report here only the analysis of the PI model in \cref{tab:valid_scores_PI}. See \cref{sec:appendix_NLI_cross_evaluation} for the NLI model results.

\cref{tab:valid_scores_PI} reports large differences in performance when the tokens are removed according to our scores, compared to random removal. When one third of tokens from both sentences is discarded, the PI model performance decreases by 2.5\%, whereas a random removal causes a 15.1\% drop (\cref{tab:valid_scores_PI}, 4th row and 4th column). The models differ most when a half of the tokens are removed, resulting in a difference in accuracy of 18.3\% compared to the baseline (\cref{tab:valid_scores_PI}, 6th row and 6th column).
Examining performance up to the removal of 20\% of tokens, the difference between the random and importance-based word removal are not so significant, probably because of the inherent robustness of the PI model which mitigates the effect of the (random) removal of some important tokens. On the other hand, removing half of the tokens is bound to have strong effects on the accuracy of the PI model, especially when some important words are removed (in the random deletion scheme); this is where removing words based on their low importance score makes the largest difference. At higher dropping rates, the random and the importance-based method tend to remove increasing portions of similar words, and their scores tend to converge (in the limiting case of 100\% removal, both strategies have exactly the same effect).
Overall, these results confirm that our method is robust with respect to the choice of the initial task and that it delivers scores that actually reflect word importance.

\subsection{Important Words are High in the Tree}
\label{sec:syntax_trees}

\begin{table}[t]
    \centering
    \small
    \scalebox{0.9}{%
    \begin{tabular}{l|rr|rr|r}
               & \multicolumn{2}{c|}{\bf NLI Model} & \multicolumn{2}{c|}{\bf PI Model} &       \\\hline\hline
    \bf Depth      & \bf Avg    & \bf Std      & \bf Avg    & \bf Std       & \bf Count     \\\hline
    1	&	\bf 0.52	&	0.35	&	\bf 0.64	&	0.31	&	9424	\\
    2	&	\bf 0.36	&	0.36	&	\bf 0.53	&	0.39	&	27330	\\
    3	&	\bf 0.23	&	0.31	&	\bf 0.40	&	0.35	&	26331	\\
    4	&	0.22	&	0.31	&	0.33	&	0.36	&	7183	\\
    5	&	0.22	&	0.30	&	0.35	&	0.35	&	1816	\\
    \end{tabular}}
    \caption{Importance scores of tokens for each depth in syntactic trees. Stat. significant differences between the current and next row are bolded ($p<0.01$).}
    \label{tab:syntactic_tree}
\end{table}

Linguistic theories differ in ways of defining dependency relations between words. One established approach is motivated by the `reducibility' of sentences \cite{lopatkova2005modeling}, i.e.\ gradual removal of words while preserving the grammatical correctness of the sentence.
In this section, we study how such relationships are also observable in attributions. We collected syntactic trees of input sentences with UDPipe \citep{straka-2018-udpipe},\footnote{\url{https://lindat.mff.cuni.cz/services/udpipe/}} which reflect syntactic properties of the UD format \cite{11234/1-4758}.\footnote{UD favors relations between content words, function words are systematically leaves in the tree. However, having function words as leaves better matches our perspective of information importance flow, unlike in \citet{gerdes-etal-2018-sud}.} When processing the trees, we discard punctuation and compute the average score of all tokens for every depth level in the syntactic trees. We display the first 5 depth levels in \cref{tab:syntactic_tree}.

We can see tokens closer to the root in the syntactic tree obtain higher scores on average. We measure the correlation between scores and tree levels, resulting in -0.31 Spearman coefficient for the NLI model and -0.24 for the PI model. Negative coefficients correctly reflect the tendency of the scores to decrease in lower tree levels. It thus appears that attributions are well correlated with word positions in syntactic trees, revealing a relationship between semantic importance and syntactic position.

\subsection{Dependency Relations}
\label{sec:dependency_relations}

\def\under#1{\underline{\smash{#1}}}
\begin{table}[t]
\centering
\small
\setlength{\tabcolsep}{3pt}
\scalebox{0.9}{%
\begin{tabular}{l|rr|rr|r}
& \multicolumn{2}{c|}{\bf NLI Model} & \multicolumn{2}{c|}{\bf PI Model}       \\\hline\hline
\bf Dependency Relation & \bf Avg & \bf Std & \bf Avg & \bf Std & \bf Count \\\hline
det, case, cop, cc, punct, mark & -0.50 & 0.37 & -0.37 & 0.49 & 34034  \\\hline
advcl, acl, xcomp & 0.11 & 0.43 & 0.06 & 0.38 & 2789 \\\hline
nsubj & -0.22 & 0.45 & 0.06 & 0.39 & 9323 \\
punct & -0.53 & 0.35 & 0.24 & 0.35 & 8148 \\
compound & 0.07 & 0.46 & -0.04 & 0.35 & 2437 \\
\end{tabular}}
\caption{The average and standard deviation of significance scores, and the count of aggregated dependency relations in syntactic trees.}
\label{tab:dependency_relation}
\end{table}

We additionally analyze dependency relations occurring more than 100 times by computing the score difference between child and parent nodes, and averaging them for each dependency type. In \cref{tab:dependency_relation}, we depict relations which have noteworthy properties with respect to significance scores (the full picture is in \cref{sec:appendix_dependence}). Negative scores denote a decrease of word significance from a parent to its child. We make the following observations.

The first row of the table illustrates dependencies that have no or very limited contribution to the overall meaning of the sentence. Looking at the corresponding importance scores, we observe that they are consistently negative, which is in line with our understanding of these dependencies.

The second row corresponds to cases of clausal relationships. We see an increase in importance scores. This can be explained since the dependents in these relationships are often heads of a clause, and thus contribute, probably more than their governor, to the sentence meaning. It shows models' ability to detect some deep syntactic connections.

The last block represents relations that are not consistent across the models. Nominal Subject is judged less important in the NLI model than in the PI model. As mentioned in \cref{sec:pos}, Punctuation differs similarly. Elements of Compound are preferred in different orders depending on the model. On the other hand, all other relation types are consistent: Ranking each type of dependency relation based on its average score and calculating correlation across our models results in 0.73 Spearman coefficient. This reveals a strong correlation between importance and syntactic roles.

\section{Conclusion}
In this paper, we have proposed a novel method to compute word importance scores using attribution methods, aiming to explain the decisions of models trained for semantic tasks. We have shown these scores have desired and meaningful properties: Content words are more important, scores are context-dependent and robust with respect to the underlying semantic task. In our future work, we intend to exploit these word importance scores in various downstream applications.
\section*{Limitations}

Our method of identifying important words requires a dataset for a semantic task (in our case NLI or PI), which limits its applicability. This requirement also prevents us from generalizing our observations too broadly: we tested our method only on one high-resource language where both dependency parsers and NLI / PI datasets are available. Our analysis also lacks the comparison to other indicators of word significance.

\section*{Acknowledgements}

The work has been partially supported by the grants 272323 of the Grant Agency of Charles University, 19-26934X (NEUREM3) of  the  Czech  Science Foundation and SVV project number 260~698. A part of this work has been done at Laboratoire Interdisciplinaire des Sciences du Numérique (LISN) in Orsay, France.

\bibliography{anthology,custom}
\bibliographystyle{acl_natbib}

\appendix

\begin{table*}[t]
    \centering
    \small
    \setlength{\tabcolsep}{3pt}
    \scalebox{0.9}{%
    \begin{tabular}{c|ccccccccccc}
              & \multicolumn{10}{c}{\bf NLI Model performance}    \\\hline\hline
    & \bf 0\%    & \bf 10\%       & \bf 20\%  & \bf 30\%    & \bf 40\%       & \bf 50\% & \bf 60\% & \bf 70\% & \bf 80\% & \bf 90\% & \bf 100\%    \\\hline
    0\% & \it 78.4$\uparrow$0.0 & \it 78.1$\uparrow$0.5 &  77.9$\uparrow$3.1 &  77.1$\uparrow$5.5 &  75.8$\uparrow$8.1 &  72.0$\uparrow$8.9 &  68.6$\uparrow$8.5 &  63.7$\uparrow$7.3 &  55.9$\uparrow$5.1 &  46.8$\uparrow$3.1 & \it 33.5$\uparrow$0.0 \\
    10\% & \it 78.4$\uparrow$1.4 &  78.3$\uparrow$1.8 &  78.1$\uparrow$4.7 &  77.2$\uparrow$6.5 &  75.7$\uparrow$8.9 &  72.1$\uparrow$9.4 &  68.6$\uparrow$9.0 &  63.6$\uparrow$7.6 &  55.8$\uparrow$5.5 &  46.6$\uparrow$3.0 & \it 33.6$\uparrow$0.1 \\
    20\% &  78.0$\uparrow$4.1 &  77.8$\uparrow$4.3 &  77.7$\uparrow$6.4 &  77.1$\uparrow$8.4 &  75.4$\uparrow$9.7 &  72.0$\uparrow$10.2 &  68.2$\uparrow$9.6 &  63.6$\uparrow$8.3 &  55.7$\uparrow$5.7 &  46.7$\uparrow$3.8 & \it 33.5$\uparrow$0.3 \\
    30\% &  77.3$\uparrow$6.5 &  77.2$\uparrow$6.6 &  77.0$\uparrow$8.8 &  76.7$\uparrow$10.3 &  74.9$\uparrow$11.2 &  71.3$\uparrow$11.8 &  68.1$\uparrow$11.1 &  63.2$\uparrow$8.9 &  55.7$\uparrow$6.7 &  46.6$\uparrow$3.9 & \it 33.4$\uparrow$0.6 \\
    40\% &  76.1$\uparrow$8.3 &  76.0$\uparrow$8.6 &  75.9$\uparrow$9.9 &  75.3$\uparrow$11.0 &  74.0$\uparrow$11.9 &  71.1$\uparrow$12.5 &  67.4$\uparrow$10.9 &  63.1$\uparrow$9.5 &  55.7$\uparrow$7.7 &  47.1$\uparrow$5.1 & \it 33.5$\uparrow$0.2 \\
    50\% &  72.8$\uparrow$8.6 &  72.7$\uparrow$8.6 &  73.1$\uparrow$10.2 &  72.4$\uparrow$10.2 &  71.5$\uparrow$11.3 &  69.3$\uparrow$12.6 &  66.7$\uparrow$12.4 &  62.4$\uparrow$10.2 &  55.5$\uparrow$8.1 &  46.4$\uparrow$4.5 & \it 33.5$\downarrow$0.2 \\
    60\% &  68.7$\uparrow$6.7 &  68.5$\uparrow$6.9 &  68.9$\uparrow$7.9 &  68.6$\uparrow$9.1 &  67.7$\uparrow$9.5 &  66.1$\uparrow$10.6 &  64.3$\uparrow$10.8 &  60.8$\uparrow$9.9 &  54.0$\uparrow$6.9 &  45.9$\uparrow$3.8 & \it 33.4$\downarrow$0.2 \\
    70\% &  63.2$\uparrow$5.3 &  63.0$\uparrow$5.2 &  63.5$\uparrow$6.3 &  62.9$\uparrow$6.0 &  62.2$\uparrow$7.0 &  61.3$\uparrow$7.8 &  60.2$\uparrow$8.8 &  58.1$\uparrow$9.5 &  52.5$\uparrow$6.2 &  45.1$\uparrow$3.4 & \it 33.4$\uparrow$0.1 \\
    80\% &  57.4$\uparrow$3.6 &  57.3$\uparrow$3.6 &  57.7$\uparrow$3.7 &  57.2$\uparrow$3.3 &  57.1$\uparrow$4.1 &  56.5$\uparrow$5.4 &  55.1$\uparrow$4.9 &  53.8$\uparrow$6.0 &  50.3$\uparrow$4.9 &  44.9$\uparrow$3.7 & \it 33.4$\downarrow$0.0 \\
    90\% &  52.5$\uparrow$2.1 &  52.4$\uparrow$2.1 &  52.9$\uparrow$2.6 &  52.8$\uparrow$2.1 &  52.4$\uparrow$2.3 &  51.9$\uparrow$2.9 &  51.2$\uparrow$2.7 &  49.9$\uparrow$2.5 &  47.6$\uparrow$3.2 &  43.5$\uparrow$3.2 & \it 33.7$\uparrow$0.4 \\
    100\% & \it 42.8$\uparrow$0.0 & \it 42.8$\uparrow$0.1 & \it 43.5$\uparrow$0.1 & \it 43.8$\uparrow$0.2 & \it 44.5$\uparrow$0.5 & \it 44.7$\uparrow$0.5 & \it 45.1$\uparrow$0.4 & \it 44.2$\downarrow$0.8 & \it 43.1$\downarrow$0.1 & \it 40.2$\uparrow$0.3 & \it 33.8$\uparrow$0.0 \\
    \multicolumn{11}{c}{} \\
    \end{tabular}}
    \caption{The accuracy of the NLI model when a given percentage of the least important input tokens are removed from the premise (rows) or hypothesis (columns) according to the PI model's weights. The description of the cell content is the same as in \cref{tab:dependency_relation}.}
    \label{tab:valid_scores_NLI}
\end{table*}

\begin{table*}[t]
\centering
\small
\setlength{\tabcolsep}{4pt}
\scalebox{0.9}{%
\begin{tabular}{l|rr|rr|r|l}
& \multicolumn{2}{c|}{\bf NLI Model} & \multicolumn{2}{c|}{\bf PI Model}       \\\hline\hline
\bf Dep. Rel. & \bf Avg & \bf Std & \bf Avg & \bf Std & \bf Count & \bf Description \\\hline
cop & -0.74 & 0.30 & -0.74 & 0.27 & 1623 & Copula, e.g. John \textit{is} the best \under{dancer}; Bill \textit{is} \under{honest} \\
case & -0.55 & 0.35 & -0.54 & 0.30 & 7651 & Case Marking, e.g. the \under{Chair} \textit{'s} office; the office \textit{of} the \under{Chair} \\
punct & -0.53 & 0.35 & 0.24 & 0.35 & 8148 & Punctuation, e.g. \under{Go} home \textit{!} \\
aux & -0.51 & 0.34 & -0.67 & 0.27 & 4622 & Auxiliary, e.g. John \textit{has} \under{died}; he \textit{should} \under{leave} \\
cc & -0.48 & 0.32 & -0.74 & 0.23 & 707 & Coordinating Conjunction, e.g. \textit{and} \under{yellow} \\
det & -0.45 & 0.38 & -0.55 & 0.38 & 14801 & Determiner, e.g. \textit{the} \under{man} \\
mark & -0.39 & 0.34 & -0.48 & 0.31 & 1104 & Marker, e.g. \textit{before}; \textit{after}; \textit{with}; \textit{without} \\
nsubj & -0.22 & 0.45 & 0.06 & 0.39 & 9323 & Nominal Subject, e.g. \textit{John} \under{won} \\
nummod & -0.10 & 0.37 & -0.02 & 0.38 & 1269 & Numeric Modifier, e.g. \textit{forty} \under{dollars}, \textit{3} \under{sheep} \\
nmod & -0.06 & 0.52 & -0.13 & 0.42 & 3153 & Nominal Modifier, e.g. the \under{office} of the \textit{Chair} \\
advmod & -0.01 & 0.51 & -0.01 & 0.41 & 1299 & Adverbial Modifier, e.g. \textit{genetically} \under{modified}, \textit{less} \under{often} \\
advcl & 0.05 & 0.43 & 0.05 & 0.33 & 857 & Adverbial Clause Modifier, e.g. if you \under{know} who did it, you should \textit{say} it \\
compound & 0.07 & 0.46 & -0.04 & 0.35 & 2437 & Compound, e.g. \textit{phone} \under{book}; \textit{ice} \under{cream} \\
conj & 0.10 & 0.41 & 0.03 & 0.28 & 742 & Conjunct, e.g. \under{big} and \textit{yellow} \\
acl & 0.11 & 0.43 & 0.04 & 0.41 & 1367 & Adnominal Clause), e.g. the \under{issues} as he \textit{sees} them; a simple \under{way} to \textit{get} \\
amod & 0.11 & 0.42 & -0.01 & 0.32 & 2974 & Adjectival Modifier, e.g. \textit{big} \under{boat} \\
obl & 0.16 & 0.47 & 0.09 & 0.33 & 5002 & Oblique Nominal, e.g. last \textit{night}, I \under{swam} in the \textit{pool} \\
xcomp & 0.21 & 0.41 & 0.12 & 0.38 & 565 & Open Clausal Complement, e.g. I \under{started} to \textit{work} \\
obj & 0.25 & 0.44 & 0.12 & 0.36 & 4377 & Object, e.g. she \under{got} a \textit{gift} \\
\end{tabular}}
\caption{The average and standard deviation of significance scores, and the count and a short description of each dependency relation between a \under{parent} and \textit{child} node in the syntactic tree.}
\label{tab:dependency_relation_full}
\end{table*}

\section{Training}

\subsection{Underlying Models \label{sec:appendix_models}}

\begin{table}
    \centering
    \small
    \scalebox{0.9}{%
    \begin{tabular}{l|rrr|r}
               & \multicolumn{3}{c|}{\bf Training}       \\\hline\hline
               & \bf Entail.    & \bf Neutral. & \bf Contra.       & \bf ~~~~All     \\\hline
    \textsc{snli}       & 183k       & 183k    & 183k          & 549k  \\
    \textsc{qnli}       & 52k        & 52k     & -               & 105k  \\
    \textsc{multi\_nli} & 131k       & 131k    & 131k          & 393k  \\\hline
    All        & 366k       & 366k    & 315k          & 1047k \\
    \multicolumn{5}{c}{} \\
    \end{tabular}}
    
    \scalebox{0.9}{%
    \begin{tabular}{p{40px}|rrr|r}
               & \multicolumn{3}{c|}{\bf Validating}       \\\hline\hline
               & \bf Entail.    & \bf Neutral. & \bf Contra.       & \bf ~~~~All     \\\hline
    \textsc{snli}    & 3.3k       & 3.3k    & 3.3k          & 10k  \\
    \end{tabular}}
    
    \caption{The number of samples in training (top) and validation (bottom) data for the NLI model.}
    \label{tab:nli_datasets}
\end{table}

\paragraph{Implementation} Language modeling often treats the input of semantic classification tasks as a one-sequence input, even for tasks involving multiple sentences on the input side \citep{devlin-etal-2019-bert,lewis-etal-2020-bart,lan2019albert}. However, processing two sentences as one irremediably compounds their hidden representations. As we wish to separate representations of single sentences, we resort to a custom implementation based on the Transformers architecture \citep{vaswani2017attention}, which comprises two encoders (6 layers, 8 att. heads, 1024 feed forward net. size, 512 emb. size) sharing their weights, one for each input sentence. Following Sentence-BERT \cite{reimers-gurevych-2019-sentence}, we computed the mean of the encoder output sentence representations $u$ and $v$, and concatenated them to an additional $|u-v|$ term. This was passed to a linear layer for performing the final classification. We implemented models in \texttt{fairseq} \citep{ott-etal-2019-fairseq}.\footnote{\url{https://github.com/facebookresearch/fairseq}}

\paragraph{Datasets} The NLI model was trained on \textsc{snli} \citep{bowman-etal-2015-large}\footnote{\url{https://huggingface.co/datasets/snli}}, \textsc{multi\_nli} \citep{williams-etal-2018-broad}\footnote{\url{https://huggingface.co/datasets/multi_nli}} and \textsc{qnli} \cite{rajpurkar-etal-2016-squad}\footnote{\url{https://huggingface.co/datasets/glue\#qnli}} datasets. Since \textsc{qnli} uses a binary scheme (`entailment' or `non-entailment'), we interpret `non-entailment' as a neutral relationship. \cref{tab:nli_datasets} describes the NLI training and validation data. The PI model was trained on \textsc{Quora} Question Pairs\footnote{\url{https://huggingface.co/datasets/quora}} and \textsc{paws} \citep{zhang-etal-2019-paws}\footnote{\url{https://huggingface.co/datasets/paws}} datasets. We swapped a random half of sentences in the data to ensure the equivalence of both sides of the data. \cref{tab:paraphrase_datasets} displays the PI training and validating data.

\paragraph{Training} We trained both models using an adaptive learning rate optimizer ($\alpha = 3\times10^{-4}$, $\beta_1 = 0.9$, $\beta_2 = 0.98$) \citep{kingma2015adam} and a inverse square root scheduler with 500 warm-up updates. We trained with 64k maximum batch tokens over 6 epochs with 0.1 dropout regulation. We trained on an NVIDIA A40 GPU using half-precision floating-point format FP16, which took less than 2 hours for both models. The PI model and NLI model achieve 85.1\% and 78.4\% accuracy on corresponding validating sets, respectively. We consider this performance sufficient given limitations put on the architecture choice.

\begin{table}
    \centering
    \small
    \scalebox{0.9}{%
    \begin{tabular}{l|rr|r}
               & \multicolumn{2}{c|}{\bf Training}       \\\hline\hline
               & \bf Paraphrase    & \bf Non Paraphrase       & \bf ~~~~All     \\\hline
    \textsc{quora}       & 146k       & 248k          & 394k  \\
    \textsc{paws} & 25k       & 55k          & 80k  \\\hline
    All        & 171k     & 303k & 474k \\
    \multicolumn{4}{c}{} \\
    \end{tabular}}
    
    \scalebox{0.9}{%
    \begin{tabular}{l|rr|r}
               & \multicolumn{2}{c|}{\bf    Validating}       \\\hline\hline
               & \bf Paraphrase    & \bf Non Paraphrase       & \bf ~~~~All     \\\hline
    \textsc{quora}    & 3.4k       & 3.4k          & 6.8k  \\
    \end{tabular}}
    
    \caption{The number of samples in training (top) and validation (bottom) data for the PI model.}
    \label{tab:paraphrase_datasets}
\end{table}

\subsection{Interpreter}
\label{sec:appendix_interpreter}

\paragraph{Implementation} We use the attribution method introduced by \citet{de-cao-etal-2020-decisions}. Assuming $L$ layers for the NLI encoder, the interpreter model contains $L + 1$ classifiers. Each classifier is a single-hidden-layer MLP, which inputs hidden states and predicts binary probabilities whether to keep or discard input tokens. The implementation details closely follow the original work.

\paragraph{Training} We trained on the first 50k samples of the corresponding underlying model's training data, using a learning rate $\alpha = 3\times10^{-5}$ and a divergence constrain $D_* < 0.1$. The number of training samples and the rest of hyper-parameters follow the original work.
We trained over 4 epochs with a batch size of 64.

\section{Cross-Task Evaluation}
\label{sec:appendix_NLI_cross_evaluation}

The performance of the NLI model in the cross-task evaluation, compared to the baseline model, is displayed in \cref{tab:valid_scores_NLI}.

\section{Dependency Relations}
\label{sec:appendix_dependence}

We examined all dependency relations with a frequency greater than 100 by computing the score difference between child and parent nodes, and averaging them for each every dependency type. Results are in \cref{tab:dependency_relation_full}.

\end{document}